\documentclass[twocolumn,10pt]{asme2e}
\usepackage{graphicx}
\usepackage{amsmath}

\confshortname{DSCC2019}
\conffullname{the ASME 2019 \\
              Dynamic Systems and Control Conference}

\confdate{9-11}
\confmonth{October}
\confyear{2019}
\confcity{Park City, Utah}
\confcountry{USA}

\papernum{DSCC2019-9097}

\title{BEHAVIOR IDENTIFICATION AND PREDICTION FOR A PROBABILISTIC RISK FRAMEWORK}

\author{Jasprit Singh Gill, {\tensfb Pierluigi Pisu}, {\tensfb Venkat N. Krovi}, {\tensfb Matthias J. Schmid\thanks{corresponding author}}     
    \affiliation{Department of Automotive Engineering\\
	Clemson University International Center of Automotive Research\\
	Greenville, South Carolina, 29607\\
	jasprig@g.clemson.edu, pisup1@asme.org, vkrovi@clemson.edu, schmidm@clemson.edu
    }
}

\begin{document}

\maketitle    

\begin{abstract}
{\it Operation in a real world traffic requires autonomous vehicles to be able to plan their motion in complex environments (multiple moving participants). Planning through such environment requires the right search space to be provided for the trajectory or maneuver planners so that the safest motion for the ego vehicle can be identified. Given the current states of the environment and its participants, analyzing the risks based on the predicted trajectories of all the traffic participants provides the necessary search space for the planning of motion. This paper provides a fresh taxonomy of safety / risks that an autonomous vehicle should be able to handle while navigating through traffic. It provides a reference system architecture that needs to be implemented as well as describes a novel way of identifying and predicting the behaviors of the traffic participants using classic Multi Model Adaptive Estimation (MMAE). Preliminary simulation results of the implemented model are included.}
\end{abstract}



\section*{INTRODUCTION}

Deployment of highly autonomous vehicles on the road requires them to be able to plan their motion in complex environment(environments with multiple maneuvering participants). Planning problem in any environment can basically be considered as an optimization problem with following steps: (i) identifying the possible states (i.e. search space) that the vehicle can attain while staying within safety constraints; (ii) performing an optimized search through this search space using an objective function (cost function) that represents the identified constraints of the motion; and (iii) provide  a reference to the motion execution modules of the vehicle. Depending on the different factors considered while generating a search space, the different levels of planning are involved in navigation.

\begin{enumerate}
\item Path planning:
A path is a collection of poses that a vehicle has followed in the past or might be following in a finite future \cite{gillNavigationEval2019}. It usually consists of states like longitudinal position, lateral position and orientation in case of 2D navigation in a road environment but without any information of time. Path planning is determining the future path of the vehicle given its present states, available map and mission level goals. States of the other participants of the traffic environment are usually not considered in this. 

\item Trajectory planning:
Trajectory is a collection of position, orientation, velocity and/or higher derivative states of a vehicle marked with time stamps. A path can be extracted from a trajectory, but trajectory cannot be recreated from path due to lack of time information\cite{gillNavigationEval2019}. Trajectory planning determines the future trajectory of the vehicle given its present states, map, goal and may also involve states of the participants of the environment.

\item Maneuver planning:
A maneuver is a collection of motion sequences that a vehicle executes in order to achieve a local level mission objective, e.g. lane change, lane merging, left turn, right turn. As seen in the literature\cite{gindele2010probabilistic, xie2018vehicle,lefevre2015intention,lefevre2014survey}, the terms maneuver and behavior will be used interchangeably in this document. Maneuvers can typically be described by trajectories since they involve specifying stopping times, velocities and possibly accelerations at different spatiotemporal points in the future. Maneuver planning can either be planning to identify a maneuver or it can encompass trajectory planning. 
\end{enumerate}

Navigation through a complex traffic environment requires planning to be performed at the trajectory level or maneuver level. Given the current states of the environment and its participants, analyzing the risks provides the necessary search space for such planners. Risk analysis can be decomposed into following 3 steps\cite{lefevre2014survey,de2014collision,lefevre2015intention}: (i) identify states of all the traffic participants; (ii) predict their future states in a time step t+s in the future; and (iii) check the possibility of a risky event (e.g. collision) at each time step.

To assess the risks in the correct domain space for effective planning, we need to understand the needs of real-world driving scenarios. Collisions with other entities (pedestrian, bicyclists, other vehicles, stationary objects, etc.) are the primary risks that need to be avoided. However, the risks can also include violating a road traffic rule, road ethics, or even inconvenient driving conditions that can lead to an injury to the occupants or the vehicle functionality in the short or long term. For instance, the drivers actively avoid behaviors that can lead to a probable collision. Hence, identifying the behaviors and predicting the trajectories of all the traffic participants becomes a key in analyzing the risks.

This paper seeks to approach the challenges of navigation through complex traffic environments by collectively looking at the needs of the operating domain as well as the state-of-the-art of the technology from a broader perspective followed by a narrowed down identification of solutions. The contribution of this paper are as follows:

\begin{enumerate}
 \item It provides a fresh taxonomy of safety / risks that an autonomous vehicle needs to address while navigating through traffic. 
 \item It presents a reference system architecture suited for implementation in an autonomous vehicle in order to navigate through traffic successfully. Then, it discusses an existing probabilistic framework in the literature and demonstrates how it can be mapped to the architecture. 
 \item It describes a novel way of identifying and predicting the behaviors of the traffic participants using classic Multi Model Adaptive Estimation (MMAE) and demonstrating how it can be integrated into the discussed probabilistic framework. Finally, some preliminary simulation results of implemented model are described.
\end{enumerate}

The paper is organized as follows. The next section describes the classification of risk assessment based on the needs of the traffic environment. This is followed by the reference system architecture and the proposed model for behavior identification. Due to the breadth of the topics covered in this paper, the related work has been discussed in the respective sections.


\section*{RISK ASSESSMENT CLASSIFICATION}

\begin{figure}[t!]
\includegraphics[width=0.5\textwidth]{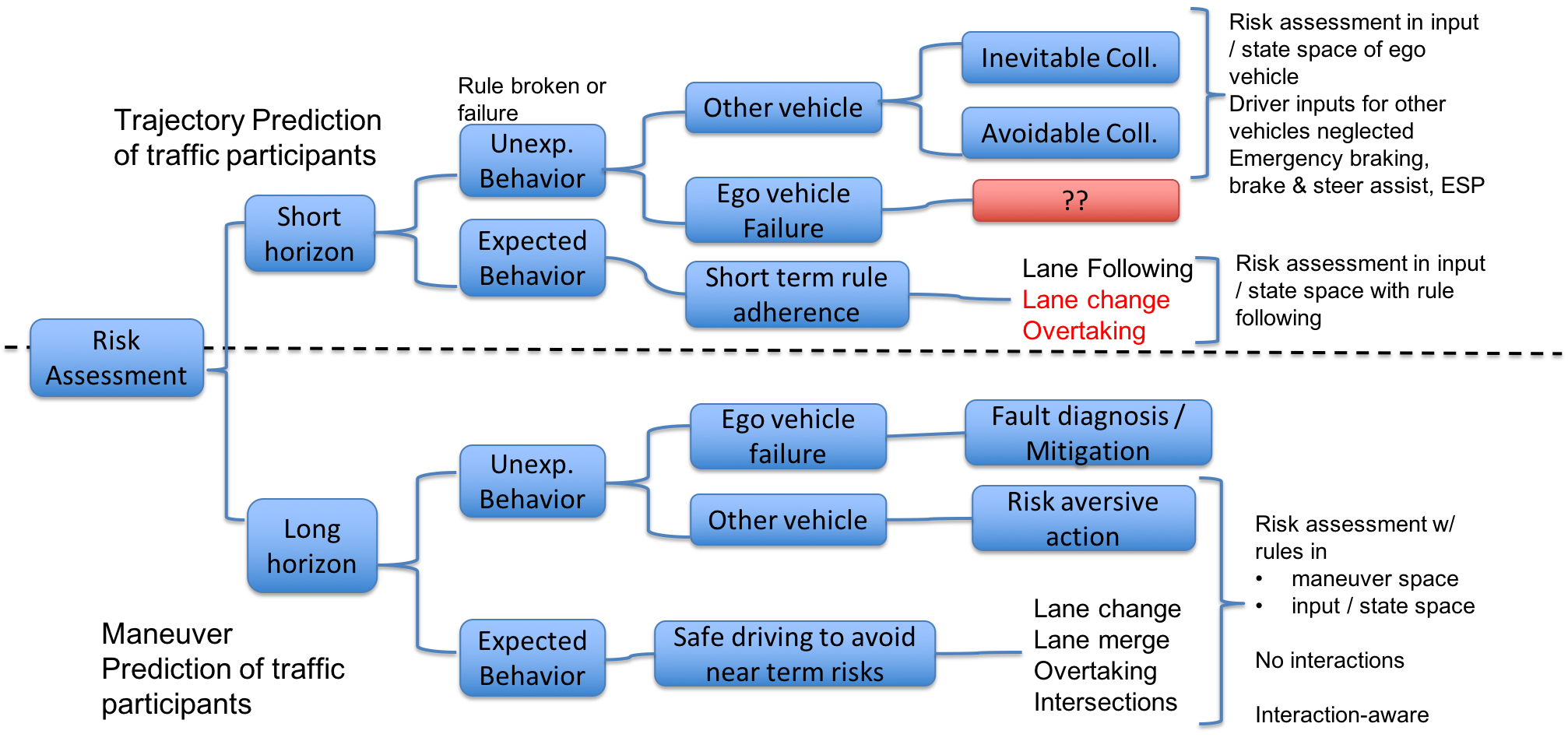}
\caption{RISK ASSESSMENT CLASSIFICATION} 
\label{fig2}
\end{figure}

At the broadest level, the risk assessment can be classified as either near term (short horizon) or	 long horizon as described in figure \ref{fig2}. When the foreseen evaluation of the risk is in the temporal vicinity of human response time, its in the near term else it can be considered long horizon. This classification is important due to one key factor in determining the risk mitigation strategy (i.e. motion planning) - whether the traffic rules should be followed or not. For near term risk prevention, the primary priority of the planners needs to be risk avoidance with or without rule following. For long horizon risk prevention, the planners need to obey the traffic rules. As a result mitigation strategies for both the categories of risks require different policy models to be followed. In both the cases, identifying whether a traffic participant performs  an expected behavior as per the traffic rules or an unexpected behavior, can further help identify a risky situation. One of the recent works in identifying such unexpected behaviors was presented in \cite{lefevre2015intention}.

\subsection*{Near term risk assessment} 
For the near-term risk assessment, in the literature, the interactions between the traffic participants are ignored. Physics based models (like constant velocity, constant acceleration, constant turning rate acceleration, constant curvature, etc.  \cite{LiTrackingDynamicSys2001}) are used for predicting the trajectories in the future as if the participants are moving independent of each other. The argument that the interactions between the vehicles can be ignored holds true only for the time horizon of a human response time, which is about 1-2 seconds.

If an unexpected behavior is observed such that the other traffic participant (due to a rule breaking or a failure) is at fault, the risk analysis needs to determine if the collision is inevitable or avoidable. Metrics such as time-to-collision, time-to-react, distance-to-collision, etc. can be utilized for this. The example situations in this category are: a participant cutting across while the ego car drives straight, or an on-coming vehicle crossing the divider and coming into the lane.  If the collision is inevitable, then the risk analysis module needs to be able to provide the assessment of impact so that a behavior that minimizes the damage can be planned while the system is preparing for a collision. Most of the emergency advanced driver assistance systems (ADAS) fall in this category. For the case in which the collision is avoidable, the risk analysis module should be able to provide the assessment of the situation so that an evasive behavior (braking only, steering only or a combination of braking and steering) with least possible consequences can be planned. Whether the collision is avoidable or inevitable, the planning in this case is done with a lower priority attributed to the rule following and higher priority attributed to the maneuver that leads to the least possible consequence.

If an unexpected behavior is observed for an ego vehicle, which is likely due to a failure in some vehicle module, then the fault diagnosis and fault mitigation routines need to take over as typical models may fail in this case. This research area is out of scope of the presented work. 

In case where all the participants are behaving as they should, and no rule is broken, the main objective of the risk assessment module is to help short term rule adherence for the ego vehicle and analyze the trajectories that may lead to dangerous situations. Driver assistance systems such as lane keeping, adaptive lane following (adaptive cruise control), etc fall under this category. 

\subsubsection*{Related work}
Most of the ADAS systems address risks in this category. Lane merge assist systems that are on the horizon also address such risks, however a lane merge maneuver isn't always a short term maneuver. Collision-based risk prediction has been an active topic for researchers for quite some time now with a large body of literature available. The popular approach in such cases is to ignore the driver inputs, make assumptions about constant velocity, acceleration, steering angle or steering rate, predict the trajectories of the traffic participants based on this, and then check for an event like time-to-collision, distance-to-collision\cite{lefevre2014survey}, time to closest approach, etc. to determine possible conflicts. Some of the notable works that implement computationally effective methods to determine possible conflicts are discussed here. Campos et. al. \cite{de2014collision} provide the 3 step procedure for the threat assessment for collision avoidance at intersections. The three step procedure is: (i) use unscented transform to predict the trajectories using constant turn rate acceleration (CTRA) model; (ii) define some geometric areas (collision zones approximated as rectangles) on the vehicles for calculating the time-to-collision (TTC) and distance-to-collision (DTC) and then use bivariate normal distribution integral approximation (Drezner, 1978\cite{drezner1978computation}) to find the probability of collision. (iii) employ reachability analysis for assessment of threat. This publication also gives the values of covariance parameters from experiments, which are typically challenging to determine. Batz et al \cite{batz2009recognition} use unscented transform for predicting trajectories. Here, covariance ellipses are utilized in calculating the area representing the position of the vehicles, adding uncertainty of orientation and then applying Minkowski's sum operator to it for every vehicle. Then for every vehicle pair, the approach calculates the minimum distance in the short time in the future, if this distance falls below certain threshold it flags a dangerous situation. However, these methods are usable only for short term prediction as most of them ignore driver intent, which is the key uncertainty for long horizon risk assessment.

\subsection*{Long horizon risk assessment}
When the risks being analyzed are spatially or temporally far enough to take risk averse actions to gracefully handle the situation, it comes under this category. The assessment in this case is more for a possible risk rather than an event, although this is as important as the latter. The models in this case can assume the traffic participants to be either interaction-aware or non-interactive, however for complete risk assessment former needs to be considered. Irrespective of whether an unexpected behavior was observed or not, the rule adherence in the planning needs to be ensured in this case. 

If an unexpected behavior on part of another participant is observed, the risk assessment module analyzes different behaviors to help plan a risk-averse maneuver that gracefully handles the situation. The risk assessment in this case can either be based on the possible maneuvers that the ego vehicle can possibly make. Or it can simply be on the input space of the vehicle after compounding all the risk assessment from different possible maneuvers \cite{tuprints6790}. Example in this case would be a vehicle broken down in the middle of the road. Or an intersection scenario where the ego vehicle comes across another vehicle from the perpendicular road stopped in the middle of an intersection while the signal is green. The graceful behavior in such cases would be for the ego car to either brake to a complete stop or changing the lanes to drive around the vehicle. 

If no unexpected behaviors are observed for the traffic participants, then the risk assessment in this case will be safety analysis and efficiency analysis of driving maneuvers such that ones that can lead to a possible near-term risk can be avoided. Examples of such situations can be when a car is parked on the shoulder. Typical behavior in this case of the vehicles driving in the adjoining lane is either to slow down or change the lanes. Another example would be in a multiple lane intersection with merge ins, where other vehicles are merging in, behaviors of the vehicle in the adjoining lane can be changing the lane so as to avoid any possible conflicts.

For the long term risks, the steps that need to be followed for risk assessment are:
1.	Model and predict the possible expected behaviors of the traffic participants.
2.	Analyze the possible risk of every possible behavior of the ego vehicle, against the possible predicted behaviors of the traffic participants.
3.	Once the traffic participant executes the behavior, the module should be able to identify the behavior.
4.	If an unexpected maneuver was executed, the module should be able to differentiate it from the expected ones.

\subsubsection*{Related work}

The frameworks and the works addressing the challenges in this area are relatively recent. Gindele et. al. \cite{gindele2010probabilistic} presented one of the first frameworks that considers vehicle interaction as well as the driver behavior for prediction. This framework will be discussed in detail in the section Probabilistic Framework. Lefevre et al. \cite{lefevre2015intention} presented an extension of this framework for reasoning for collision risk at a semantic level by differentiating expected behaviors of the drivers from the unexpected behaviors. The framework was demonstrated by applying to the road intersections with interactions. Among the most recent works, Damerow \cite{tuprints6790} presents a situation-based risk evaluation and behavior planning framework for highly automated driving. Analysis of risks is performed using prototypical predicted trajectories of the traffic participants and the result of the analysis is a proposed risk map that basically a search space for the behavior planner. The framework is demonstrated for parallel driving and intersection traffic scenes. Prototypical trajectories based approaches however implicitly assume the availability of the high definition digital maps for effective operation. Further, the framework is decoupled from the environment perception module and doesn't reuse the information about uncertainty already available from it. However, the prediction sub-module of our reference architecture described in the next section, is inspired from this framework.


\section*{SYSTEM ARCHITECTURE FOR HIGHLY AUTONOMOUS DRIVING}

\begin{figure}[t!]
\includegraphics[width=0.5\textwidth]{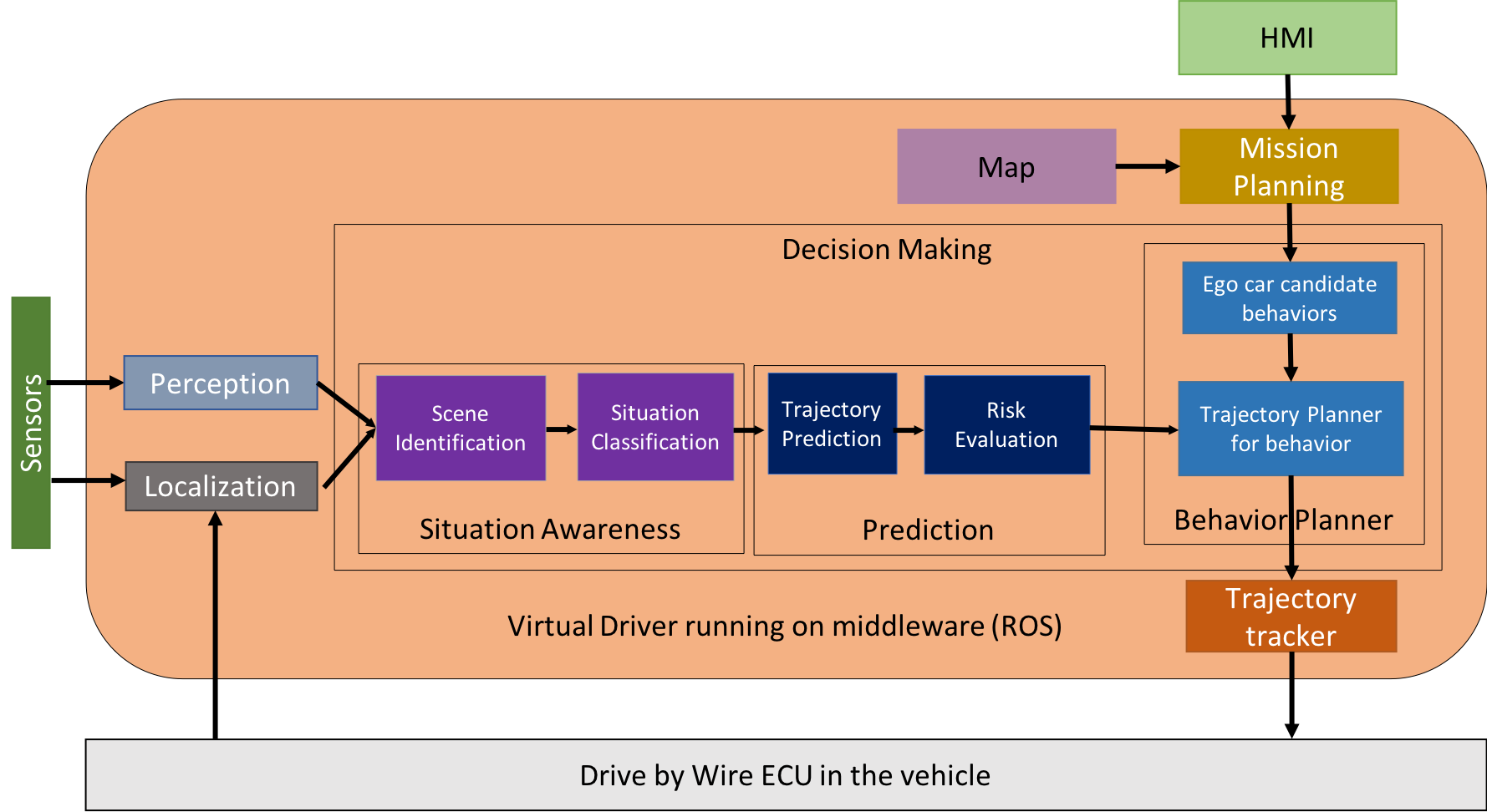}
\caption{AUTONOMOUS VEHICLE ARCHITECTURE} 
\label{SW_Architecture}
\end{figure}

Figure \ref{SW_Architecture} provides the overall software architecture envisioned for a highly autonomous driving vehicle. The different modules in it are described below. 

Localization: This module uses proprioceptive and exteroceptive sensors to determine the detailed pose (position and orientation) of the ego-vehicle, including the lane and offset from the center lane, in the operating environment. Typical proprioceptive sensors are inertia measurement unit (IMU) and wheel speed sensor, whereas commonly used exteroceptive sensors are camera, LiDAR and GPS. 

Perception (Environment perception): Uses exteroceptive sensors like cameras, lidar and radar to detect stationary and dynamic obstacles that are around ego vehicles and, classifies them and tracks them to estimate their kinematic states. It also separates quasi-stationary features in the environment (obstacles) from permanent features that define the environment (like landmarks, lane information, traffic signs, for e.g.). 

Decision making module: The information about the ego-vehicle pose, the states of the other traffic participants and traffic conditions (signals, road geometry) determined from the localization and perception module are fed into the decision-making module. This module can be further divided into situational awareness (SA) and prediction sub-modules. 

Situational awareness sub-module: This module uses the information from perception module to identify traffic scenarios.  Traffic scenarios can be thought of as a high level semantic representation of the traffic environment, like signalized intersection, non-signalized intersection, lane following, round about, etc. Each traffic scenario can be further divided into different situations depending on the larger objective of the ego vehicle and the current states of the other traffic participants. For e.g., in an intersection where the ego vehicle intends to turn left, the situation can vary depending on whether there is another participant present in an oncoming lane that is going straight. This identification is important as it defines the traffic policies (like right of the way) that apply to all the traffic participants in the situation. The identification of an unexpected behavior by a traffic participant needs to happen in this module. After identifying the traffic scene, based on the current states of all the traffic participants, this sub-module is also responsible for identifying the current situation of the scenario. 

Prediction sub-module: Based on the current situation, prediction module is responsible for determining the possible situations into which the current situation can evolve, depending on the possible behaviors all the traffic participants may perform. For each possible situation, the module then needs to make some assumptions about every traffic participant and then forward simulate their trajectories for a predefined look ahead. These predicted trajectories of all traffic participants can then used to analyze the safety or risks involved, considering the traffic policies in place, against every possible maneuver that the ego-vehicle can take. The metric for this conflict can be based on lane occupancy, possible future collision, time to collision, distance to collision, etc. The outcome of a risk analysis is a search space that the planning modules then use to plan the trajectory of the ego vehicle.

Mission planning module: This module uses the mission level parameters like current position, goal position and digital map information to plan the most optimal route for the vehicle. The planned route doesn't consider traffic or lane level information. Further it only consists of waypoint or path information and not trajectory information. This module is responsible in determining when the assigned mission has been accomplished. 

Behavior planning module: This module takes the route information from the mission planning module, and the risk/safety information from the decision-making module to identify the safest behavior that the ego-vehicle should perform specifying the trajectory to realize the behavior. Based on this a reference trajectory is generated and provided to the trajectory tracker module.

Trajectory tracker: Based on the reference trajectory it receives from the behavior planner and the current state of the vehicle, this module executes the behavior in a smoothest possible way. 

As can be seen, for an effective navigation, the situational awareness, prediction and behavior planner work together in the decision making. The success of behavior planner in navigating safely through a dynamic environment is dependent on the capability of the prediction module to identify the risks. The prediction module in turn depends on the situational awareness module to identify the right traffic policy applicable based on the situation and then analyze the risk.  As a result, unlike near term risk planning, where these modules are decoupled from each other, for a comprehensive analysis of long-term risks, there is a significant information flow and interaction between the three modules requiring a probabilistic framework that meets this requirement.
 
\section*{PROBABILISTIC FRAMEWORK}
Currently, the most comprehensive framework available for prediction is provided by Gindele et. al. \cite{gindele2010probabilistic}. The framework introduces the concept of context and establishes the relationships between context, situations and behaviors. This section will briefly cover the filter equations and explain how it maps to our reference architecture. For more details the reader is referred to the work \cite{gindele2010probabilistic}.

The joint probability density function of the dynamic bayesian network our work adopts from is given by:
\begin{multline}
P(X, X^-, C, S, B, B^-, T, T^-, Z) = \\
P(X^-)P(B^-)P(T^-)P(X|T^-)\\
P(C|X)P(S|C)P(B|B^-, S)P(T|T^-, B, C, X)P(Z|X)
\label{eq_joint_density}
\end{multline}
The resulting filter equation can be given by
\begin{multline}
P(X, C, B, T, U|z) \propto \\
P(z|X)\int_{B^-, T^-}P(X, C, S, B, T|B^-, T^-)P(B^-; T^-)
\label{eq_filter_update}
\end{multline}
where, for N traffic participants in the scene, 
X: is the vector containing the states of all the traffic participants. 
$$X = (X_1\  X_2\  X_3\ ...\ X_N)^T $$
C: Vector containing the context, which is a list of features such as distances from the surrounding traffic participants, that help define a situation
$$C = (C_1\  C_2\  C_3\ ...\ C_N)^T $$
S: Vector containing the situations of all the traffic participants
$$S = (S_1\  S_2\  S_3\ ...\ S_N)^T $$
B: Vector containing the possible behaviors of all the traffic participants for a particular situation.
$$X = (B_1\  B_2\  B_3\ ...\ B_N)^T $$
T: Vector containing the trajectories that realize the behaviors for all the traffic participants.
$$T = (T_1\  T_2\  T_3\ ...\ T_N)^T $$
Z: Measurement vector for every observed vehicle
$$Z = (Z_1\  Z_2\  Z_3\ ...\ Z_N)^T $$
If we compare this with the system architecture presented in figure \ref{SW_Architecture}, the probability density function (pdf) of the prior ($X^-$) is provided collectively by the perception and the localization modules. The scene identification sub-module in the situational awareness block, uses the available digital maps and / or the prior information to identify the traffic scene. The situation classification sub-module then utilizes this information, derives the context information using helper functions (for details about helper functions refer to \cite{gindele2010probabilistic,agamennoni2012}) and uses them to provide the conditional pdf of the situation ($P(S|C)$). Based on the traffic policies defined within it, the trajectory prediction sub-module of the prediction module, provides a conditional pdf of the possible behaviors that all the traffic participants may perform ($P(B|B^-, S)$). It then identifies the likely trajectories that the traffic participants might execute to realize the behaviors ($P(T|T^-, B)$). The risk evaluation sub-module uses these predicted trajectories of the traffic participants to predict possible risks as per the identified metrics.

The focus of this study is the behavior identification of the traffic participants and their trajectory prediction. Gindele et al employed multiple motion models for different behaviors of the participants, however they realized the framework using particle filters. The difference in our approach lies in the utilization of Multiple Model Adaptive Estimation (MMAE)\cite{crassidis2004optimal} for modeling the behaviors. This has two benefits: first, a variant of MMAE called interactive multiple model (IMM) filter\cite{LiTrackingDynamicSys2001, crassidis2004optimal, kaempchen2004imm} is already one of the prevailing techniques in the tracking implementations of perception modules. Hence, our approach helps integrating into the existing frameworks of perception. Second, the approach is less computationally intensive than particle filters.  

\section*{BEHAVIOR IDENTIFICATION MODEL WITH MULTIPLE MODEL FILTERS}
The presence of both discrete as well as continuous states in the autonomous vehicle framework described above makes it a hybrid system. MMAE-based approaches are the prevailing techniques in the hybrid state estimation. For a comprehensive review of MMAE based approaches the reader is suggested to refer to\cite{LiTrackingMultimodel2005,crassidis2004optimal}. These works also provide a classification of the MMAE into 3 generations: a) classic MMAE, b) the Interacting Multi Model (IMM) approach; and c) the variable structure MMAE approach. Of the above, the IMM aproach is adopted the most as it has proven to be superior to the classic MMAE \cite{LiTrackingMultimodel2005} and due to their simplicity compared to variable structure approaches. Both classic MMAE and IMM run a bank of kalman filters, each with a different motion or measurement models, and then provide a combined state estimate calculated from the weighted state estimate of every filter. The IMM approach however, has an additional step called interaction step in which the most recent estimates from all single-model filters are mixed according to their predicted probabilities and then set as initial estimates for their next cycles. It is this interaction step that makes the execution of IMM filter banks dependent on each other in between every cycle. As a result, the models in IMM can run truly parallel to each other only within a prediction step, but not across multiple prediction cycles. In classic MMAE, on the other hand, all single model filters in its bank run independent of each other, and hence can run truly parallel across all cycles. This enables utilization of the recent advances in the parallel computing technology. 

One related recent study by Xie et al also uses a multi model approach for prediction. The authors employ an IMM to differentiate between long term and near term trajectory prediction for the same maneuver. In contrast, our study differs in two ways: first, We use the classic MMAE for trajectory prediction. Second, each single-model filter corresponds to a different maneuver. With the recent advances in computing and sensor technologies, and due to a different set of requirements for prediction applications as compared to tracking applications, a careful re-evaluation of MMAE is necessary. For the scope of this work, the MMAE approach is exemplary for only one observed vehicle. Expanding to multiple vehicle will be addressed in future work.

The classic MMAE has three stages \cite{crassidis2004optimal}: 

Model specific filtering: The equations for prediction and update for each single-model filter follow the extended kalman filter (EKF) \cite{crassidis2004optimal}. The different model assumptions for each filter will be discussed in the next section.

Model probability update: This stage determines the likelihood of the measurements for every single-model filter estimate, and then determines the associated weights for each filter. For M filters, initial weight for every filter $w_0^{(j)} = 1/M$ for j = 1, 2, ... M.
\begin{align}
w_k^{(j)} &= w_{k-1}^{(j)} p(\tilde{z}_k|\hat{x}_k^{-(j)})\\
p(\tilde{y}|\hat{x}_k^{-(j)}) &= \frac{1}{\left [ det\left ( 2\pi E_k^{-(j)} \right ) \right ]^{1/2}}e^{\left \{ -1/2\ e_k^{-(j)T} (E_k^{-(j)})^{-1} e_k^{-(j)}\right \}}\\
w_k^{(j)} &\leftarrow  \frac{w_k^{(j)}}{\sum_{j=1}^{M}w_k^{(j)}}
\end{align}
%

Combination: The estimates and covariances from all single-model filters are combined with their corresponding weights to provide a combined estimate. 
\begin{align}
\hat{x}_k^{+} &= \sum_{j=1}^{M} w_k^{(j)}\hat{x}_k^{+(j)}\\
P_k^{+} &= \sum_{j=1}^{M} w_k^{(j)} \left [\left (\hat{x}_k^{+(j)}-\hat{x}_k^+ \right ) \left ( \hat{x}_k^{+(j)}-\hat{x}_k^+ \right )^T + P_k^{+(j)}  \right ]
\end{align}
\subsection*{Motion models and measurement model}
In the presented study, MMAE is applied to model three maneuvers: straight motion, lane change to the left and lane change to the right. Constant velocity model is used to for straight motions. Left and right lane changes are modeled using sinusoidal functions. A bank of three filters runs in parallel: one linear kalman filter for straight motions and two extended kalman filters (EKF) for the lane changes. The states of the traffic participants are defined as $X=(x, v_x, y, v_y)$, where x and y define the position of the vehicle in the road coordinate frame, $v_x$ and $v_y$ are longitudinal and lateral velocities of the vehicle.

\subsubsection*{Model equations for straight motion:}
For the straight motion, the discrete time state transition function for a single step $k+1$ with sampling time $T_s$ results as,
\begin{equation}
\begin{bmatrix}
x_{k+1}\\ 
v_{x(k+1)}\\ 
y_{k+1}\\
v_{y(k+1)}
\end{bmatrix} = \begin{bmatrix}
1 & Ts & 0 & 0 \\ 
0 & 1 & 0 & 0 \\ 
0 & 0 & 1 & Ts \\ 
0 & 0 & 0 & 1
\end{bmatrix} \begin{bmatrix}
x_k\\ 
v_{x(k)}\\ 
y_k\\ 
v_{y(k)}
\end{bmatrix}
\end{equation}

\subsubsection*{Model equations for lane change maneuvers:}
For the right lane change with the lane width $w_L$, length of the lane change maneuver L and longitudinal distance of the vehicle from the start of the maneuver $\Delta x$, the sinusoidal motion can be expressed as:
\begin{equation}
y\left ( \Delta x \right ) = \frac{w_L}{2}cos\left ( \frac{\pi}{L}\Delta x \right )
\label{eqn_lane_change}
\end{equation}
For simplicity, we assume that at initial condition the vehicle is at the start of the maneuver with a constant velocity. The discrete time equations of motion can then be approximated as:
\begin{align}
x_{k+1} &= x_k + v_{x k}T_s
\label{eqn_lane_change_x}\\
v_{x(k+1)} &= v_{x k}
\label{eqn_lane_change_vx}\\
y_{k+1} &= y_{k} + v_{y k}T_s
\label{eqn_lane_change_y}\\
v_{y(k+1)} &= -\frac{w_L\pi v_x}{2L}sin\left ( \frac{\pi}{L} x_{k+1} \right )
\label{eqn_lane_change_vy}
\end{align}
%
%
%
%
The Jacobian matrix for the right lane change maneuver yields:
\begin{equation}
\Phi  = \begin{bmatrix}
1 & T_s & 0 & 0 \\ 
0 & 1 & 0 & 0 \\ 
0 & 0 & 1 & Ts \\ 
a & b & 0 & 0 
\end{bmatrix} 
\end{equation}
where 
\begin{align}
a &= - \frac{w_L}{2}\left (\frac{\pi}{L}  \right )^{2} v_{xk} cos\left (\frac{\pi}{L} x_k  \right )
\label{eqn_a_jacobian}
\\[10pt]
b &= - \frac{w_L \pi}{2L}cos\left (\frac{\pi}{L} x_k  \right )
\label{eqn_b_jacobian}
\end{align}
Similarly, the equations for the left lane change can be described by a phase shifted equation for the right lane change:
\begin{equation}
y\left ( \Delta x \right ) = \frac{w_L}{2}cos\left ( \frac{\pi}{L}\Delta x - \pi\right )
\end{equation}
\subsubsection*{Measurement model}
Assuming a hybrid sensor fusion architecture\cite{lytrivis2009sensor}, the measurement model was chosen to be linear, and the positions of the vehicle were assumed to be observable. The equations can be described by:
\begin{equation}
\begin{bmatrix}
z_x\\ 
z_y
\end{bmatrix}
=
\begin{bmatrix}
1 & 0 & 0 & 0\\ 
0 & 0 & 1 & 0 
\end{bmatrix}
\begin{bmatrix}
x_k\\ 
v_{xk}\\ 
y_k\\ 
v_{yk}
\end{bmatrix}
\end{equation}
\section*{VEHICLE DYNAMICS MODEL}
In the second stage of simulations, a single track vehicle dynamics model described below was used to evaluate our behavior identification model. This model approximates the vehicle loads by considering only the front and rear axle loads and ignoring the roll and pitch motions, resulting in two translational and one rotational degree of freedoms. The equations of motion are given by
\begin{align}
\dot{v}_x - v_y \dot{\psi } &=  \frac{1}{m} \left (F_{x,f} cos(\delta ) + F_{x,r} - F_{y,f} sin(\delta)  \right ) = \frac{F_X}{m}\\[10pt]
\dot{v}_y + v_x \dot{\psi } &=  \frac{1}{m} \left (F_{y,f} cos(\delta ) + F_{y,r} - F_{x,f} sin(\delta)  \right ) = \frac{F_Y}{m} \\[10pt]
I_{zz}\ddot{\psi} &=  (l_f F_{y,f} cos(\delta )- l_rF_{y,r} + F_{x,f} sin(\delta)  = M_z  
\end{align}
%
%
where m is the mass of the vehicle, ${v_x}$ and ${v_y}$ are the longitudinal and lateral velocity, ${\psi}$ is the yaw angle and ${\delta}$ is the steering angle of the vehicle. $F_{x,i}$ and $F_{y,i}$, with i = f,r are the net longitudinal and lateral forces acting on the front and rear wheels of the vehicle. $l_f$ and $l_r$ are the distances of the front and rear axles from the center of gravity of the vehicle, and $I_z$ is the vehicle inertia. For more details as well as the parameter values selected for the equation above, the reader is referred to Berntorp et al. (\cite{berntorp2014models}). The measurements for the behavior model were synthetically generated by adding dela-correlated (white) noise to the truth from the vehicle model.

\section*{SIMULATIONS OF THE BEHAVIOR IDENTIFICATION MODEL}
\begin{figure}[t!]
\includegraphics[width=0.5\textwidth]{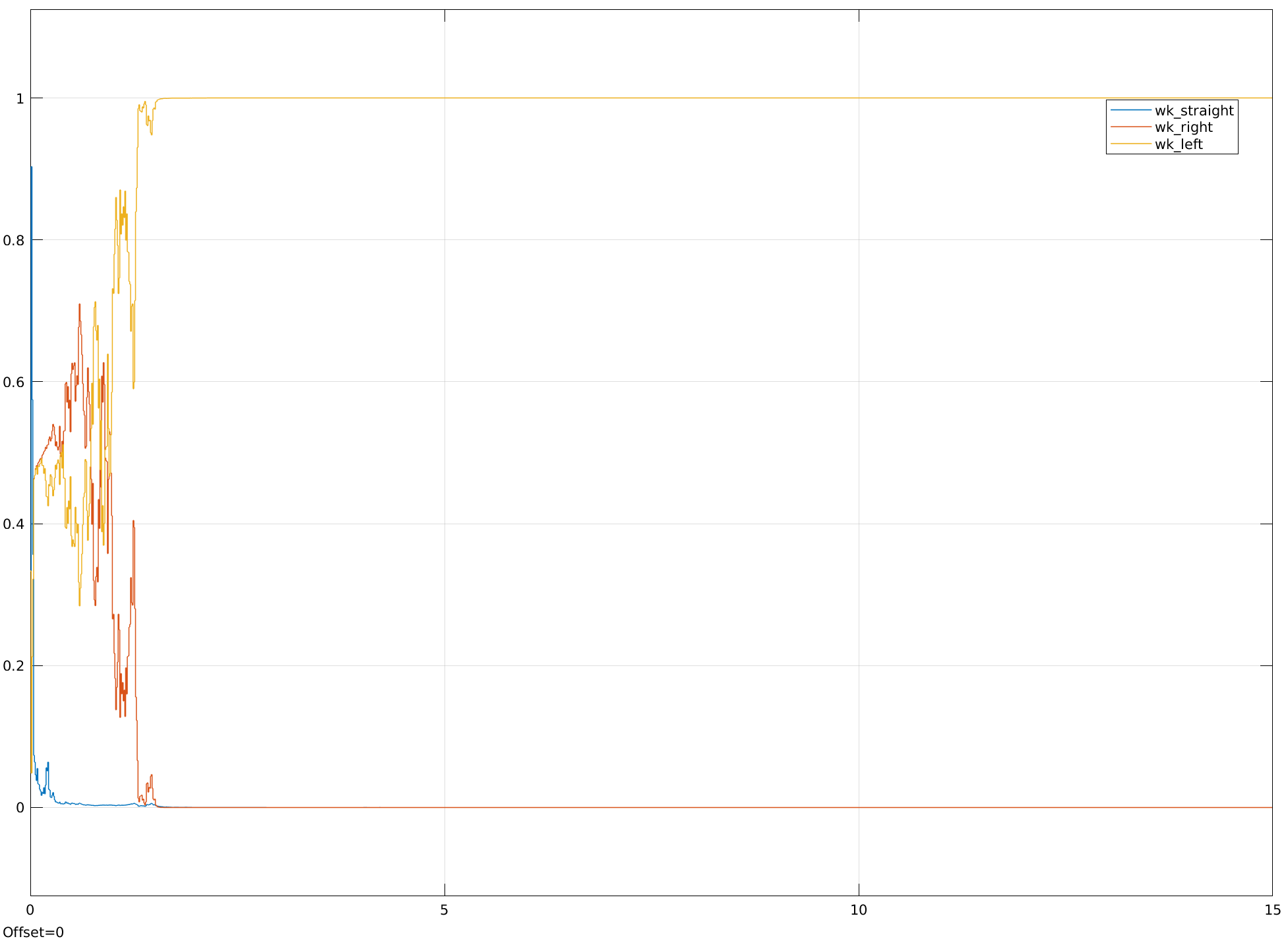}
\caption{LEFT LANE MANEUVER IDENTIFIED BY MMAE APPROACH; Y-AXIS IS WEIGHTS and X-AXIS IS TIME} 
\label{fig_left_lane1}
\end{figure}
\begin{figure}[t!]
\includegraphics[width=0.5\textwidth]{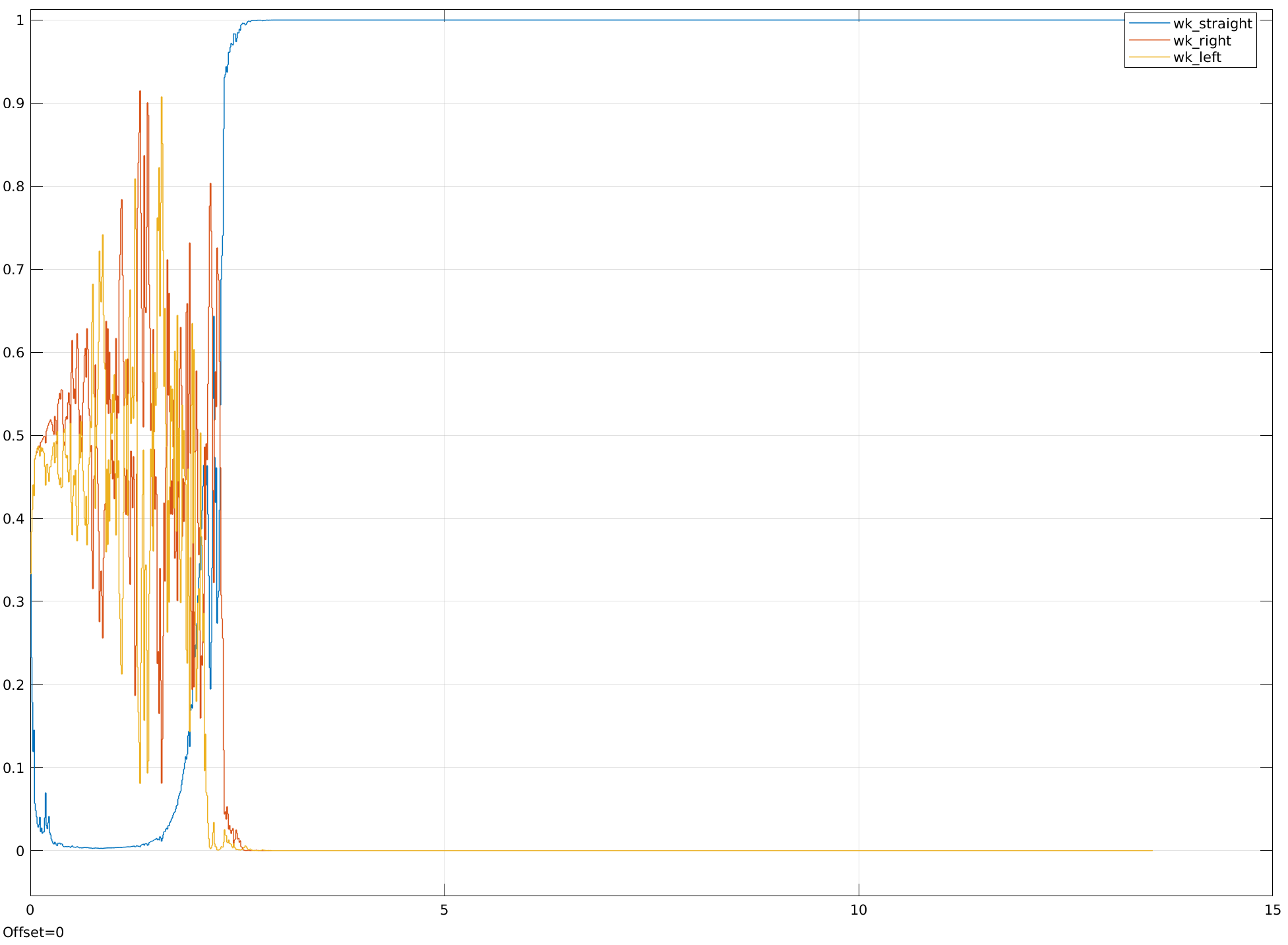}
\caption{STRAIGHT MANEUVER IDENTIFIED BY MMAE APPROACH; Y-AXIS IS WEIGHTS and X-AXIS IS TIME} 
\label{fig_straight_2}
\end{figure} 
\begin{figure}[t!]
\includegraphics[width=0.5\textwidth]{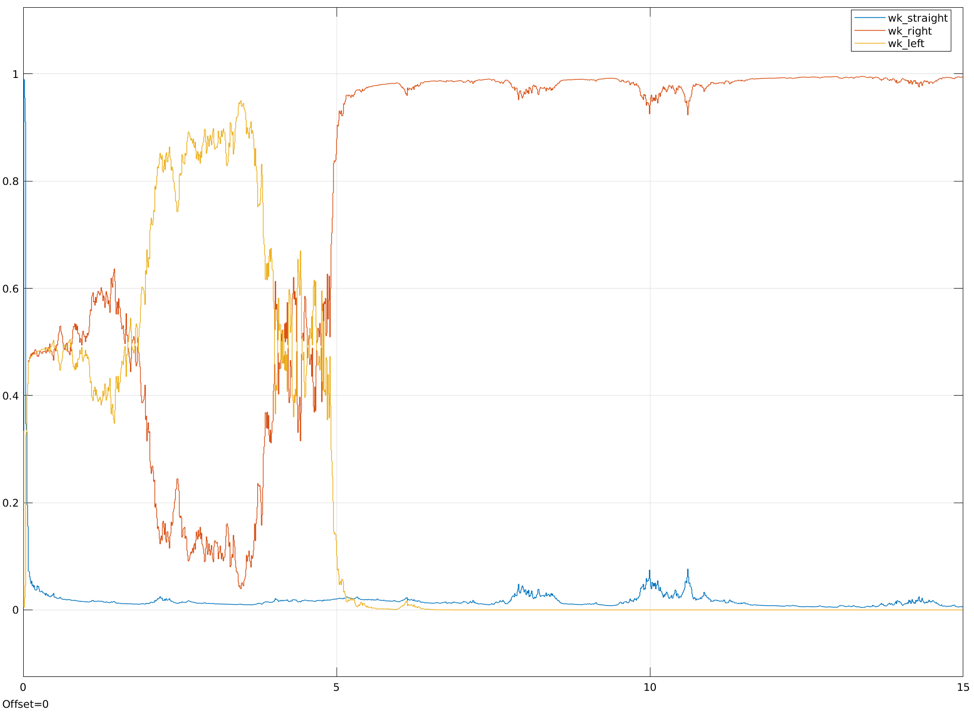}
\caption{DETECTION OF RIGHT LANE CHANGE MANEUVER FOR Q=0.1 and R=0.0025; Y-AXIS IS WEIGHTS and X-AXIS IS TIME} 
\label{G_Q_0_1_R_0_0025_right}
\end{figure}
\subsection*{Tuning of the model}
For the first stage of the simulations, the measurements for the filters were generated synthetically as described below to tune the filter. The process noise covariance matrix for all the filters was initially set to  Q = diag([0.001 0.001 0.001 0.001]). The measurement noise covariance matrix for all the filters was set to R = [0.0025 0; 0 0.0025], i.e. a standard deviation of about 0.05 m (5 cm). The measurements for positions x and y were then synthetically generated by simulating the motion models described above with additive white noise. The initial conditions for both synthetic measurements as well as the filter banks were set to x = 0 m, y = 0 m, $v_x = 10$ m/s and $v_y = 0$ m/s and their covariances $(P_0)$ set to $10^{-6}$ signifying high confidence in the initial value estimates. The maneuver length for left and right behavior models was set to 150 m. At the mentioned initial velocity, it takes about 15 secs for the maneuver to be complete.

Figure \ref{fig_left_lane1} shows that for a left lane change performed, the maneuver was detected correctly in about 1.3 seconds of initiation. The straight run on the other hand was detected in about 2.2 seconds (Figure \ref{fig_straight_2}). It needs to be noted however that these results are when the initial estimates of the observed vehicle match the initial conditions of the filter banks, and the measurement as well as the process noises are very low. Realistically however, the initial conditions of the filter may defer from the true initial conditions of the vehicle due to multiple reasons: sensor noise, point of view of the sensor, environment noise to name a few. As a result the covariance matrix of the initial condition should be set to a relatively higher value, signifying a lower confidence in the initial estimate. 

\begin{figure}[t!]
\includegraphics[width=0.5\textwidth]{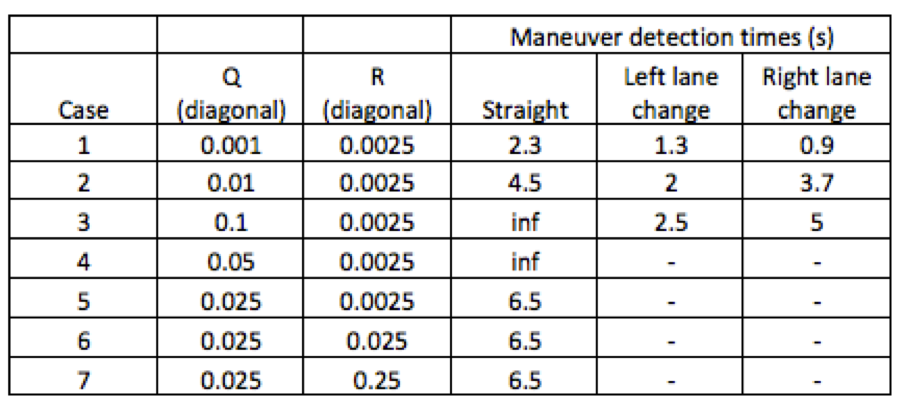}
\caption{VARIATION OF PROCESS NOISE AND MEASUREMENT NOISE ON BEHAVIOR MODEL} 
\label{Table_process_noise}
\end{figure}

Now we focus on the process noise, which should give us how much variation it can tolerate. The table in figure \ref{Table_process_noise} lists the different cases that were considered. In all these cases, the initial conditions of the filters and the actual initial conditions were identical – x0 = y0 = $v_{y0}$ = 0, $v_{x0}$ = 10. $P_0$ = 100. The process and the measurement noise was modeled as an additive white noise with same filter Q and R statistics for the synthetic measurements. As can be seen, increasing the process noise increases the detection times of the maneuver. Before detecting in each of these cases, the weights switch between the left and right lane change (LC) model filters. After a certain point (case 3), the behavior identification module is not able to identify a straight maneuver and keeps switching between left and right maneuver. For a given process noise, increasing the measurement noise had little effect on the detection times, although it did add fluctuations in the probabilistic weights that ripple down to the weighted estimates. Interestingly, there is an asymmetry in the detection times for left LC and right LC. For right LC it switches a little longer before settling down (figure \ref{G_Q_0_1_R_0_0025_right}). The cause for this is yet to be investigated.
%
\section*{Evaluation with the single track vehicle model} 
\begin{figure}[t!]
\includegraphics[width=0.5\textwidth]{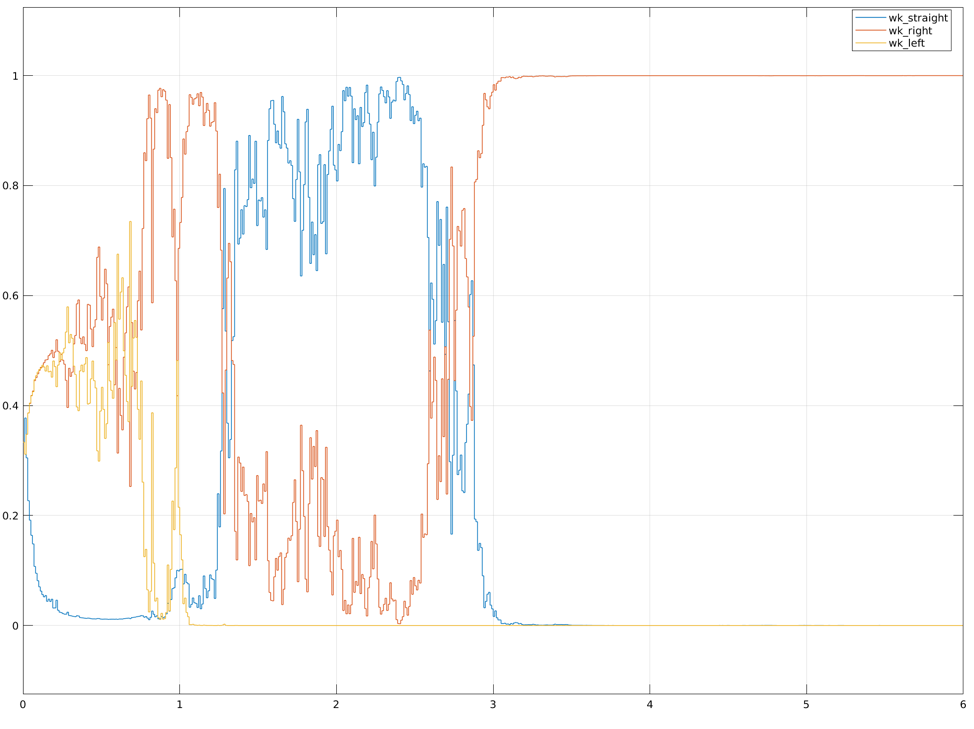}
\caption{DETECTION OF RIGHT LANE CHANGE MANEUVER WHEN Q=0.005 and R=0.0025 WITH VEHICLE MODEL; Y-AXIS IS WEIGHTS and X-AXIS IS TIME} 
\label{ST_Q_0_005_R_0_0025_right}
\end{figure}

\begin{figure}[t!]
\includegraphics[width=0.5\textwidth]{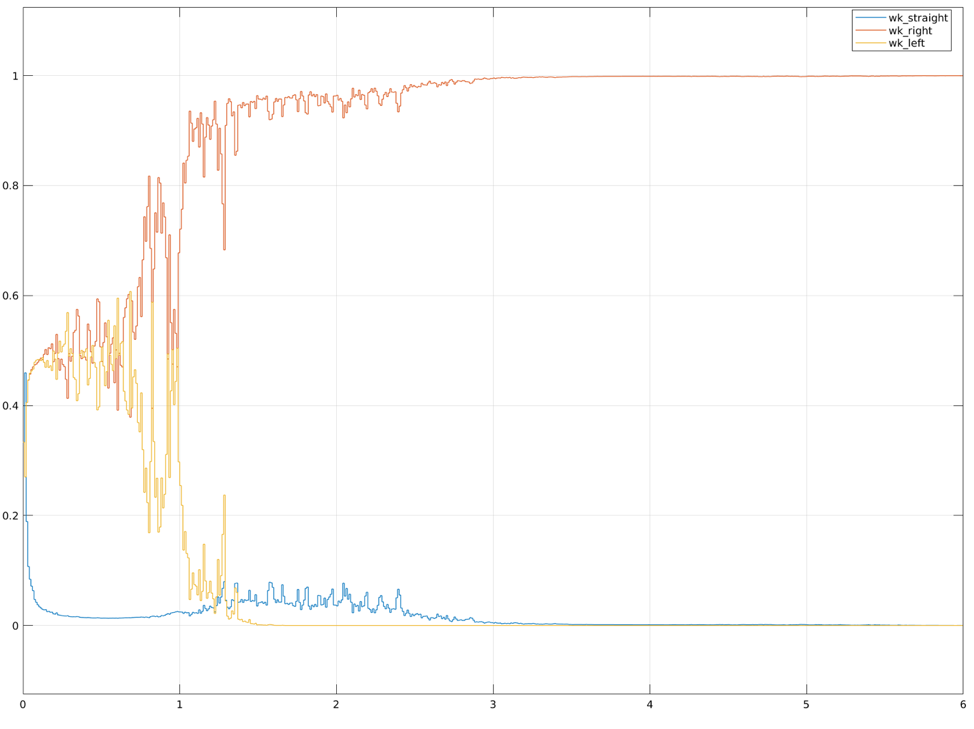}
\caption{DETECTION OF RIGHT LANE CHANG MANEUVER WHEN Q=0.025 and R=0.0025 WITH VEHICLE MODEL; Y-AXIS IS WEIGHTS and X-AXIS IS TIME} 
\label{ST_Q_0_025_R_0_0025_right}
\end{figure}

\begin{figure}[t!]
\includegraphics[width=0.5\textwidth]{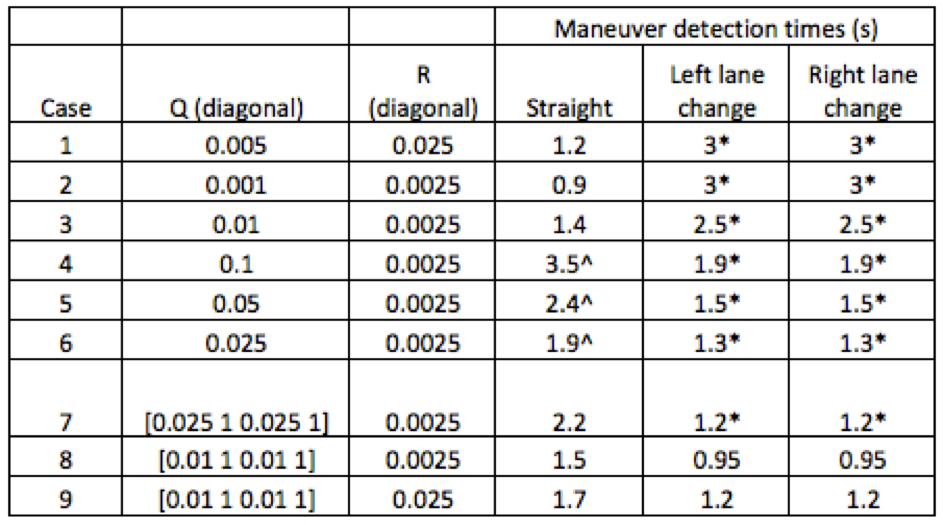}
\caption{RESULTS OF VARYING THE PROCESS AND MEASUREMENT NOISE WHEN EVALUATING WITH VEHICLE MODEL} 
\label{ST_With_Behavior_Model_table}
\end{figure}

In the second stage of simulations, the MMAE based behavior model was evaluated with the single track vehicle model as described above. The parameters chosen for the vehicle were referred from \cite{berntorp2014models}. The straight maneuver was simulated by maintaining the steering wheel angle to zero for the vehicle model. The left and right lane change maneuvers were simulated by giving a sinusoidal input of period 10 seconds to the steering wheel. The resulting maneuver length for the the lane changes was about 60 meters. Measurements were generated from this simulated vehicle model by adding brownian motion as described in the table in figure \ref{ST_With_Behavior_Model_table}. For the filters in the MMAE behavior model, the maneuver length for the right and left LC models was fixed to 60 meters. The detection times for all the maneuvers for different process and noise covariances can be seen in the table in figure \ref{ST_With_Behavior_Model_table}. For cases 1 and 2 in this table, the model confuses the lane change behaviors with the straight maneuvers for the first few seconds before eventually identifying them correctly. This can be seen in figure \ref{ST_Q_0_005_R_0_0025_right} for the case 1. The likely reason  for this is that even though the input to the steering wheel is sinusoidal, the path followed by the vehicle isn’t sinusoidal strictly. As a result, the innovations of the lane change maneuver and the straight maneuver compete with each other initially, resulting in the switching. However, this can be addressed by modeling this difference as a process noise. Figure  \ref{ST_Q_0_025_R_0_0025_right} shows the results of the right lane change for case 6 which seems to be the most optimal one for Q to be the same for all states. Realistically, the process noise will be associated primarily to the modeling errors in the velocities as they vary with different driver behaviors. Attributing more process noise to velocities compared to positions helped achieve better results in cases 8 and 9. Increasing the measurement noise had a marginal effect on the detection times, although it did add fluctuations in the weights.
\section*{CONCLUSION}
A taxonomy for different risks that an autonomous vehicle needs to address, while navigating through traffic, was presented. A reference architecture and a probabilistic framework that collectively enables addressing such risks were described. The challenge in analyzing risks in complex traffic environments exists primarily due to the uncertainty in the maneuvers drivers may execute. A novel approach for identifying and predicting the driver maneuvers using classic multi model adaptive estimation was presented and demonstrated how it can be integrated into the reference architecture. Further, preliminary results of filter tuning and its evaluation with a single track vehicle model were presented. Since there is a significant variability in the way drivers execute the maneuvers, popular approach is to model it as process noise. Hence, the results include detailed description of how the process and measurement noises affect the performance of the maneuver identification model. More comprehensive analysis is expected to be available along with the final manuscript.

\bibliographystyle{asmems4}

\begin{acknowledgment}
The authors would like to thank Srivatsan Srinivasan for implementing the vehicle dynamics model as MATLAB/Simulink source files.
\end{acknowledgment}

%
% ...

\bibliography{dscc2019_behavior_pred}

\end{document}